\documentclass[10pt,twocolumn,letterpaper]{article}

\usepackage[pagenumbers]{wacv} % To force page numbers, e.g. for an arXiv version

\newcommand{\red}[1]{{\color{red}#1}}
\newcommand{\todo}[1]{{\color{red}#1}}
\newcommand{\TODO}[1]{\textbf{\color{red}[TODO: #1]}}
\usepackage{boldline,multirow,arydshln, xcolor}
\usepackage{graphicx}

\newcommand\pad{\vspace{-0.33cm}\\} % table padding
\usepackage{amssymb}% http://ctan.org/pkg/amssymb
\usepackage{pifont}% http://ctan.org/pkg/pifont
\newcommand{\cmark}{\ding{51}}%
\usepackage{wrapfig}
\usepackage{colortbl}
\usepackage{xcolor}
\usepackage{amsmath,amsfonts,bm, bbm}

\def\gray{\color{gray}}

\usepackage{tikz}

\def\eqref#1{equation~\ref{#1}}

\def\1{\bm{1}}

\DeclareMathAlphabet{\mathsfit}{\encodingdefault}{\sfdefault}{m}{sl}
\SetMathAlphabet{\mathsfit}{bold}{\encodingdefault}{\sfdefault}{bx}{n}

\DeclareRobustCommand{\eg}{e.g.\@\xspace}
\DeclareRobustCommand{\ie}{i.e.\@\xspace}

\newcommand\codelink{https://github.com/YihongSun/FluxGraph}

\newcommand{\prev}{TubeletGraph\xspace}
\newcommand{\ours}{FluxGraph\xspace}
\newcommand{\bh}[1]{\textcolor{magenta}{BH: #1}}
\newcommand{\ys}[1]{\textcolor{cyan}{YS: #1}}

\definecolor{wacvblue}{rgb}{0.21,0.49,0.74}
\usepackage[pagebackref,breaklinks=true,colorlinks,bookmarks=false]{hyperref}

\def\wacvPaperID{125} % *** Enter the WACV Paper ID here
\def\confName{WACV}
\def\confYear{2027}

\title{Efficient Tracking and Understanding Object Transformations}

\author{Yihong Sun \\
Cornell University \\
{\tt\small yihong@cs.cornell.edu}
\and
Bharath Hariharan \\
Cornell University \\
{\tt\small bharathh@cs.cornell.edu}
}

\begin{document}
\maketitle
\begin{abstract}
Tracking objects through state transformations is essential for understanding real-world dynamics.
However, existing methods are computationally expensive.
\prev~\cite{tubeletgraph} recently showed impressive capabilities, but its inference cost ($\sim$4.4 seconds per object-frame on VOST) precludes any real-time deployment possibilities.
We observe that \prev's overhead arises from building a spatiotemporal partition of the input video: (1) entity segmentation is computed densely for every frame regardless of whether a transformation occurs, and (2) every entity in the scene is tracked, scaling cost with scene complexity rather than the number of transformations of interest.
To address both, we propose \ours, a reactive variant that uses SAM2's internal multi-mask disagreement as a lightweight trigger for transformation detection, and removes the need for tracking all entities in the given video.
\ours is $\sim$3.3$\times$ faster than \prev on VOST while improving tracking performance and preserving state graph quality.
Furthermore, we also observe consistent speedups of 3.7--10.7$\times$ across VSCOS, M$^3$-VOS, and DAVIS17 while maintaining performance.
Code is publicly available at \href{\codelink}{\codelink}.
\end{abstract}
\section{Introduction}
\label{sec:intro}

\begin{figure*}[t]
  \centering
  \includegraphics[width=\linewidth]{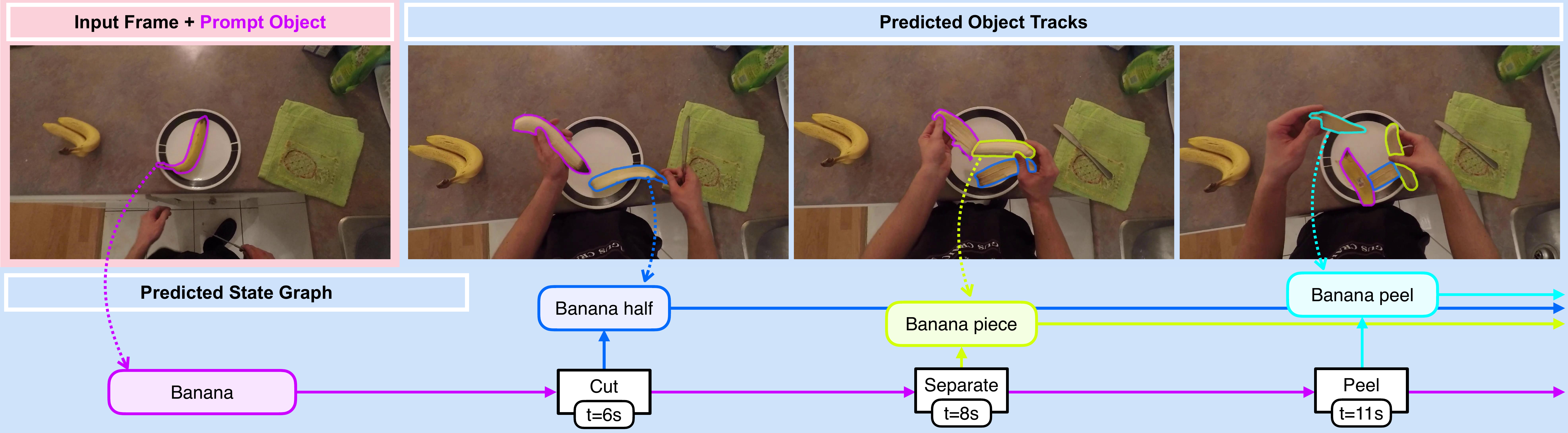}
  \caption{%
    Given a video and a prompt mask, the goal is to track the object and construct a state graph for each detected transformation and resulting effect. For this example, \ours achieves a $5.28\times$ speedup over \prev~\cite{tubeletgraph} while matching performance.
  }
  \label{fig:teaser2}
\end{figure*}

Real-world objects frequently undergo state transformations that dramatically alter their appearance: an apple sliced into pieces or a butterfly emerging from a chrysalis.
Tracking objects through these transformations is thus critical for downstream applications such as robotic manipulation~\cite{liu2024blade,sparta}, wildlife monitoring~\cite{wildlife}, and video understanding~\cite{trajtok}.
\prev~\cite{tubeletgraph} recently introduced the Track Any State task --- simultaneously tracking objects through transformations and detecting and describing the underlying state changes (Figure~\ref{fig:teaser2}).
By partitioning the input video into a set of spatiotemporal object tracks and reasoning about individual candidate tracks, \prev recovers missing objects after transformation and constructs a state graph describing each observed change.

\begin{figure}[t]
  \centering
  \includegraphics[width=\linewidth]{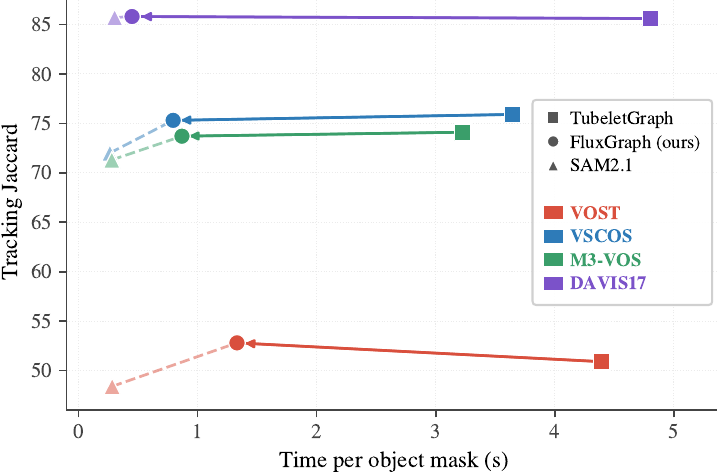}
  \caption{%
    Performance vs efficiency. Each line shows a dataset; each marker denotes a method. \textbf{\ours achieves $3.3-10.7\times$ speedups over \prev~\cite{tubeletgraph}} while matching performance.%
  }
  \label{fig:teaser}
  \vspace{-0.6cm}
\end{figure}

However, constructing a spatiotemporal partition of the input video is prohibitive.
A closer look at \prev reveals two compounding bottlenecks.
First, entity segmentation is invoked on \textit{every} frame of the video in order to account for any object that may appear in later frames, regardless of its relevance to the prompt object.
Second, every object is tracked, which makes the computational cost scale with the number of objects in the video, in addition to the video length and the number of transformations.
Both bottlenecks share a common root cause: \prev~\cite{tubeletgraph} is \textit{eager}.
It exhaustively partitions the video upfront so that any later transformation can be found through cheap retrieval, regardless of whether transformations are actually occurring.
In a complex kitchen scene with many static objects and a single transformation, \prev pays the full cost of partitioning the entire video --- nearly all of which is spent on regions and frames that are irrelevant.

This raises a natural question: \textit{is the spatiotemporal partition necessary for recovering object transformations?}
Here, we offer our \emph{key insight}: SAM2~\cite{sam2}'s internal multi-mask outputs already implicitly encode where related objects --- including products of a transformation --- are likely to emerge.
We thus need not eagerly partition the entire scene; but simply look for where SAM2 itself is uncertain.

Building on this insight, we propose \ours, an efficient variant of \prev that replaces the eager spatiotemporal partition with a reactive trigger.
Concretely, we use disagreement among SAM2's three candidate masks as a lightweight per-frame signal: when the masks agree, no transformation is suggested and all down-stream processing is skipped; when they disagree, we invoke entity segmentation to propose new candidate tracks only within the disagreement regions.
To remove potential false-positives, we track each candidate backwards and check whether it is consistent with the original prompt object.
Together, these two design choices eliminate \prev's two bottlenecks and yield significant speedups with no loss in tracking quality (Figure~\ref{fig:teaser}).
Our contributions are as follows:
\begin{enumerate}
    \item We identify and characterize the two computational bottlenecks of \prev, and propose \ours, a reactive variant that uses SAM2's multi-mask disagreement as a lightweight transformation trigger.
    \item Across four benchmarks (VOST~\cite{vost}, VSCOS~\cite{vscos}, M\textsuperscript{3}-VOS~\cite{m3vos}, DAVIS17~\cite{davis17}), \ours achieves 3.3--10.7$\times$ speedups over \prev while matching or slightly improving tracking and state graph quality.
\end{enumerate}
\section{Related Work}
\label{sec:related}

\textbf{Object Tracking.}
Object tracking~\cite{vot2016} and semi-supervised video object segmentation~\cite{davis17,davis16,ytvos} aim to segment a target object across a video given a first-frame mask.
Recent benchmarks address increasingly challenging settings, including long videos~\cite{lvos}, occlusions and crowded scenes~\cite{mose,lasot}, and complex object motion~\cite{mevis}.

Modern trackers predict consistent object tracks via appearance-driven matching, ranging from online feature finetuning~\cite{bhat2020learning, caelles2017one, maninis2018video}, template matching~\cite{hu2018videomatch, voigtlaender2019feelvos, yang2018efficient}, to memory-attention~\cite{xmem, oh2018fast, oh2019video, yang2020collaborative, yang2021associating, cutie}.
Recently, SAM2~\cite{sam2} extends Segment Anything~\cite{segmentanything} to video via memory-attention and has become the de facto base tracker.
Extensions further improve long-video robustness~\cite{sam2long}, fast-motion handling~\cite{samurai}, and distractor resolution~\cite{dam4sam}.
While these trackers achieve impressive results on standard benchmarks, they struggle when object appearance changes drastically due to state transformations, and none specifically target efficiency under transformation conditions.

\textbf{Understanding Object Transformations.}
Tracking and understanding objects undergoing transformations has been an active area of research.
VOST~\cite{vost} and VSCOS~\cite{vscos} target appearance-altering transformations from human--object interactions in ego-centric video~\cite{ego4d,ek100}, while M\textsuperscript{3}-VOS~\cite{m3vos} focuses on phase transitions (gas/liquid/solid).
ReVOS~\cite{m3vos} combines forward and reverse memory to improve tracking through transformations.
Beyond tracking, DTTO~\cite{wu2024tracking} provides box-level annotations for transforming objects, while HowToChange~\cite{oscs} and WhereToChange~\cite{spoc} localize transformation stages and spatially-progressing state change regions, respectively.
SPARTA~\cite{sparta} builds on these to enable real-world robotic manipulation tasks such as spreading, mashing, and slicing.
\prev~\cite{tubeletgraph} jointly tracks objects through transformations and constructs a state graph naming each change, but at significant computational cost.
Our work directly addresses this cost while preserving \prev's joint tracking and reasoning capability.

\textbf{Efficient Video Understanding.}
Reducing the cost of video systems is a long-standing goal, as the cost of per-frame processing scales poorly with video length and resolution.
SkipConv~\cite{habibian2021skip} and EvNet~\cite{dutson2022event} exploit temporal redundancy by skipping computation on regions with minimal changes between frames, while Delta Distillation~\cite{habibian2022delta} learns to estimate inter-frame feature deltas from a key-frame teacher.
Additionally, prior works~\cite{wu2019adaframe, korbar2019scsampler,wang2021adaptive} leverage lightweight heuristics to select the most informative frames for downstream applications and achieve large speedups.
Similarly, video-language models~\cite{ryoo2023token, ryoo2024xgen, trajtok} compress per-frame visual observation into compact spatiotemporal summaries for downstream reasoning.

\begin{figure*}[t]
  \centering
  \includegraphics[width=\linewidth]{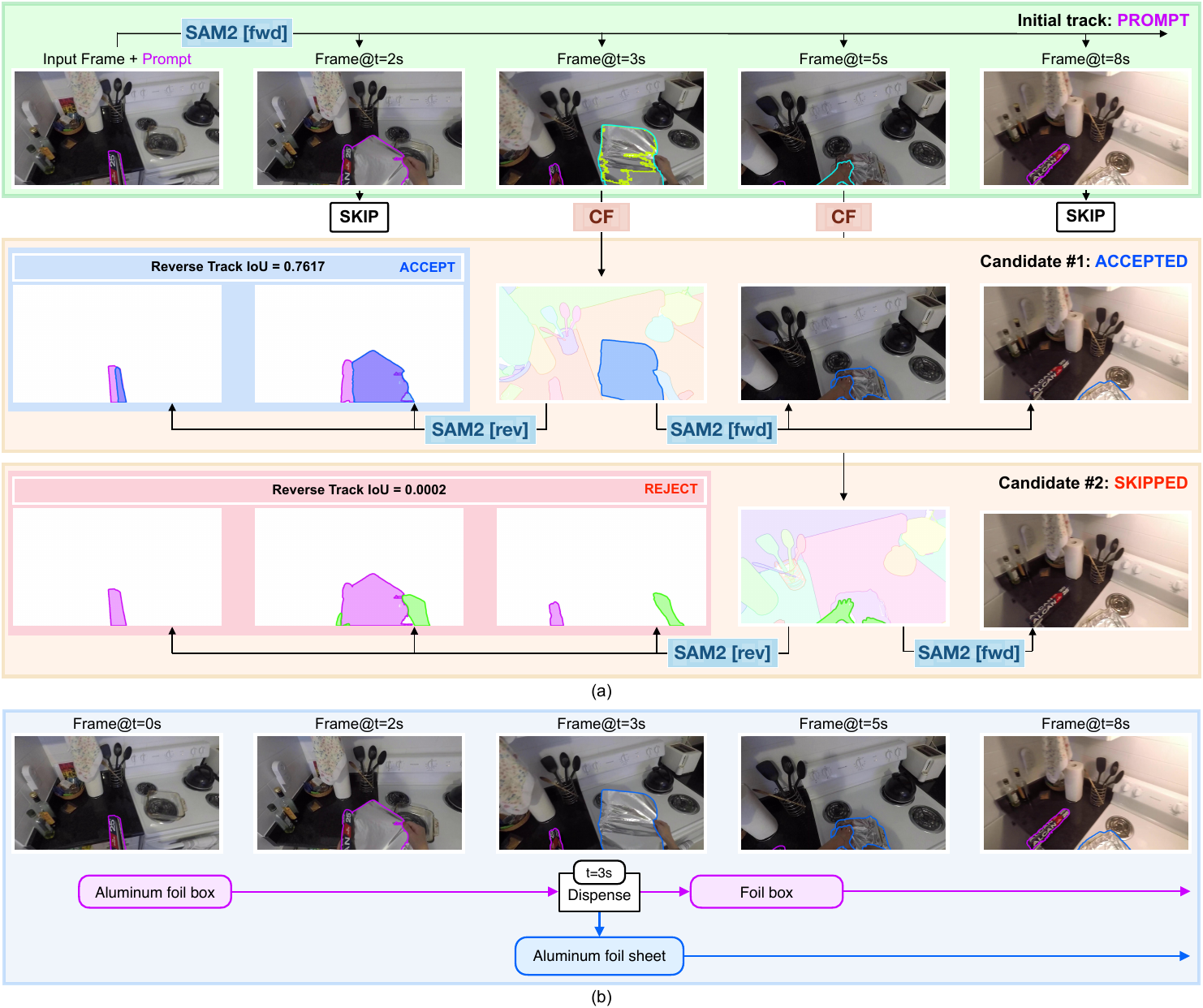}
  \caption{%
    \textbf{Overview of \ours.}
    \textbf{(a)} Given an input video and a prompt object (\textit{aluminum foil box}, magenta), \ours first propagates the prompt forward through the entire video via SAM2~\cite{sam2} to obtain the primary track $P$ (in magenta) and two alternative mask hypotheses $\{m_s^1, m_s^2\}$ (in cyan and yellow).
    Note that any overlapping region between the alternative masks and the primary track is omitted for clarity.
    Any frame where the alternatives agree with the primary is skipped (SKIP, \eg\ $t{=}2s$ and $t{=}8s$), while frames with meaningful untracked disagreement trigger an entity segmentation query via CropFormer~\cite{cropformer} (CF, \eg\ $t{=}3s$ and $t{=}5s$).
    Each entity that overlaps with the disagreement region is then added as a candidate, propagated forward, and then propagated backward in time to compare against the prompt track $P$.
    Candidate~\#1 (the emerging \textit{aluminum foil sheet}, blue) yields a high reverse-track IoU ($0.7617$) and is accepted.
    Candidate~\#2 (the unrelated \textit{actor hands}, green) yields a near-zero reverse-track IoU and is rejected as an unrelated entity.
    \textbf{(b)} The accepted candidate is merged with the prompt track, and the resulting prediction is summarized by a VLM into a state graph: the \textit{aluminum foil box} is \textit{dispensed} at $t{=}3s$, producing the persisting \textit{foil box} together with the newly emerged \textit{aluminum foil sheet.}
  }
  \label{fig:pipeline}
\end{figure*}

In the object tracking domain, efficiency efforts have largely focused on reducing the cost of SAM2~\cite{sam2}.
EfficientTAM~\cite{xiong2024efficienttam} replaces SAM2's hierarchical encoder with a plain ViT and a compressed memory module, while EdgeTAM~\cite{zhou2025edgetam} accelerates the image encoder and memory attention, respectively.
While effective, these works speed up the per-frame cost of \emph{the tracker itself}, and are largely orthogonal to the source of overhead in TubeletGraph~\cite{tubeletgraph}, which arises from invoking entity segmentation densely and tracking every scene entity.
In contrast, \ours targets to reduce the pipeline-level overhead of transformation-aware tracking by skipping expensive operations on frames and regions where minimal disagreement in SAM2's own predictions indicates that no transformation is occurring.

\section{Method}
\label{sec:method}

\ours is a reactive variant of \prev built via a single insight: SAM2's internal multi-mask outputs already encode where, and when, an object transformation is likely to introduce new components.
We start with an overview of \prev and its bottlenecks in Sec.~\ref{sec:method-recap}, and then present the two components of \ours: reactive candidate discovery via multi-mask disagreement (Sec.~\ref{sec:method-discovery}) and candidate filtering via reverse tracking (Sec.~\ref{sec:method-filtering}).

\subsection{Preliminaries}
\label{sec:method-recap}

\textbf{Task.}
Track Any State~\cite{tubeletgraph} takes as input a video $\mathcal{V} = \{I_t\}_{t=1}^{T}$ and a binary mask $\mathcal{M}_1$ for a prompt object in $I_1$.
The output is two-fold: a set of tracks $\mathcal{T} = \{T^1, \ldots, T^n\}$ that follow the prompt object through any state changes (allowing for fragmentation into multiple parts), and a set of state changes $\mathcal{S}$, where each $s \in \mathcal{S}$ is represented as a tuple $(t, T_\text{pre}, T_\text{post}, D)$ specifying the transformation time, pre- and post-condition tracks, and a textual description.
For example, in Figure~\ref{fig:pipeline}, the \textit{aluminum foil box} fragments at $t{=}3s$ into a persisting \textit{foil box} and a newly emerged \textit{aluminum foil sheet}, with the \textit{dispense} action linking the two.

\textbf{\prev pipeline.}
Given $\mathcal{M}_1$, \prev~\cite{tubeletgraph} first computes an entity segmentation $\mathcal{E}_1$ of the initial frame $I_1$ via CropFormer~\cite{cropformer}, and then tracks every entity in $\mathcal{E}_1$ forward via SAM2~\cite{sam2}.
At each subsequent frame $t$, CropFormer is invoked and any entity not yet covered by an existing track is added as an emergent candidate track.
The result is a spatiotemporal partition that covers nearly every pixel of the video.
Every emergent object is then filtered via two priors: \textit{spatial proximity}, requiring sufficient overlap with SAM2's multi-mask predictions for the prompt object, and \textit{semantic consistency}, measured via FC-CLIP~\cite{fcclip} feature similarity to the prompt.
Surviving candidates are merged with the prompt track, and a VLM is queried at each transformation event to produce the final state graph.

\textbf{Bottlenecks.}
While effective, constructing the spatiotemporal partition comes at a high computational cost.
First, entity segmentation is invoked on every frame to detect newly-emerged objects, even when no transformation is occurring.
Second, every object is tracked forward via SAM2, with cost scaling linearly in the number of entities rather than the number of transformations of interest.
On VOST~\cite{vost}, these two components together account for over 98\% of \prev's $\sim$4.4 s/obj-frame inference cost.
Reducing both motivates the central designs of \ours.

\subsection{Reactive Candidate Discovery}
\label{sec:method-discovery}

We first track the prompt object $\mathcal{M}_1$ and produce the primary prediction track $P = \{p_1, p_2, \ldots, p_T\}$ along with SAM2's two alternative mask hypotheses $\{m_s^1, m_s^2\}$ at each frame $s$ (Figure~\ref{fig:pipeline}(a), top).
These alternative masks are at the core of our reactive trigger.

\begin{figure}[t]
  \centering
  \includegraphics[width=\linewidth]{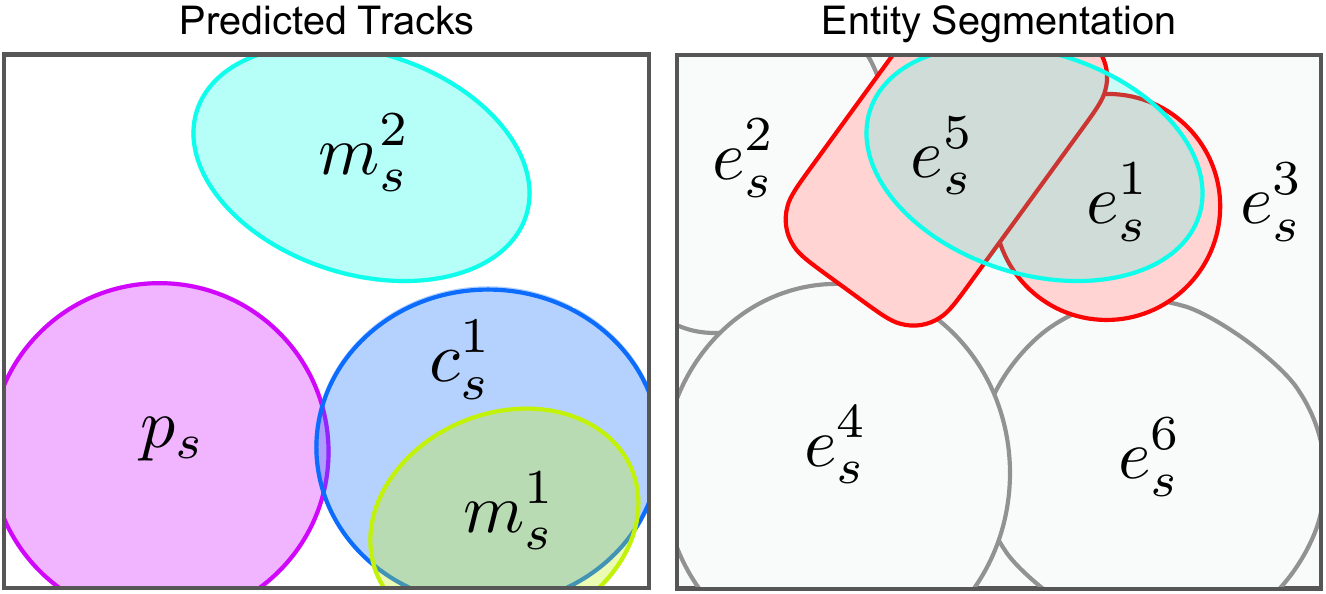}
  \caption{%
    Efficient candidate selection. Given primary mask $p_s$, alternative masks $\{m_s^1, m_s^2\}$, and tracked additional component masks from transformation $c_s^1$ (left), we perform entity segmentation (right) and select entities (in red) with sufficient coverage by alternative hypothesis $m_s^2$ and not covered by $p_s$ and $c_s^1$.
  }
  \label{fig:select}
  \vspace{-0.3cm}
\end{figure}

\textbf{Multi-mask disagreement as a transformation signal.}
A key observation we offer is that when a state transformation produces a new, separate object, SAM2 often captures it as an \emph{alternative} mask hypothesis rather than the primary prediction.
This is a natural consequence of SAM2's multi-head design and winner-takes-all training objective, which encourages the alternative heads to produce meaningful hypotheses.
Therefore, a large disagreement between $p_s$ and $\{m_s^1, m_s^2\}$ suggests that SAM2 is uncertain about where the prompt object is located --- a signal that a state transformation may have occurred at frame $s$.
This is visible at $t{=}3s$ in Figure~\ref{fig:pipeline}(a), where the alternative masks (cyan and yellow) latch onto the newly dispensed foil sheet.

This insight translates to two complementary heuristics that decide whether a frame may contain a transformation of interest.
Formally, suppose at a particular frame $s$, the prompt object is predicted with primary mask $p_s$, alternative masks $\{m_s^1, m_s^2\}$, and additional component masks from transformation $\{c_s^1, ..., c_s^n\}$, we skip all down-stream analysis if \emph{either} of the following conditions is met.

\textbf{Heuristic 1: Low alternative mask disagreement.}
If both alternative masks are nearly identical to the primary:
\begin{equation}
    \forall k \in \{1, 2\}: \quad \frac{|p_s \cap m_s^k|}{|p_s \cup m_s^k|} > \tau_\text{agree},
    \label{eq:agree}
\end{equation}
then SAM2 is confident in a single segmentation hypothesis and no new candidate region is suggested.
Shown in Figure~\ref{fig:pipeline}(a) ($t{=}2s$), the foil has not yet been dispensed and both alternative hypotheses collapse onto the primary track.

\textbf{Heuristic 2: Small residual area.}
Even when alternative masks differ from the primary, the differing region may fall entirely within already-tracked regions of previously found candidates.
Let $\mathcal{N}_s = p_s \cup c_s^1 \cup ... \cup c_s^n$ denote the union of all masks currently tracked.
If the residual area of each alternative mask after removing $\mathcal{N}_s$ is small relative to the primary mask:
\begin{equation}
    \forall k \in \{1, 2\}: \quad |m_s^k \setminus \mathcal{N}_s| < \tau_\text{res} \cdot |p_s|,
    \label{eq:res}
\end{equation}
then no meaningful untracked region is present and no new candidate region is suggested.

Consequently, we perform entity segmentation when \emph{both} conditions fail --- \ie when uncertainty in the SAM2 prediction points to an untracked region.

\textbf{Candidate region extraction.}
For each frame $s$ with entity segmentation $\mathcal{E}_s = \text{CF}(I_s)$, we identify potential candidate regions as shown in Figure~\ref{fig:select}.
For each entity $e_s^j \in \mathcal{E}_s$, we compute its coverage by the alternative masks restricted to untracked regions:
\begin{equation}
    \text{cov}(e_s^j) = \max_{k \in \{1, 2\}} \frac{|e_s^j \cap (m_s^k \setminus \mathcal{N}_s)|}{|e_s^j|},
    \label{eq:cov}
\end{equation}
and add $e_s^j$ as a candidate if $\text{cov}(e_s^j) > \tau_\text{new}$.
Intuitively, this selects entities that are covered by SAM2's own alternative hypotheses but lie outside regions already accounted for by existing tracks --- exactly where a missed transformation product would appear.
In Figure~\ref{fig:pipeline}(a), this step produces Candidate~\#1 (the foil sheet, blue) at $t{=}3s$ and Candidate~\#2 (the actor's hands, green) at $t{=}5s$.

\subsection{Candidate Filtering via Reverse Tracking}
\label{sec:method-filtering}

After identifying candidates, we propagate each one forward to the end of the video using SAM2, yielding a set of candidate tracks $\mathcal{C} = \{C^1, C^2, \ldots\}$.
However, not every candidate corresponds to a genuine product of the transformation --- some may be unrelated objects that happen to share space with the alternative multi-mask predictions, as illustrated by Candidate~\#2 in Figure~\ref{fig:pipeline}(a).

In \prev, this disambiguation is handled by combining the proximity prior with a semantic consistency check via FC-CLIP~\cite{fcclip}.
In \ours, we replace this semantic check with a purely geometric one that requires no additional model: \emph{reverse tracking}.

\textbf{Reverse tracking.}
Concretely, for each candidate track $C = \{c_s, c_{s+1}, \ldots, c_T\}$ initialized at frame $s$, we track $c_s$ \emph{backwards} in time from frame $s$ to frame $1$ (via SAM2~[rev] in Figure~\ref{fig:pipeline}(a)).
This yields a reverse track $\hat{C} = \{\hat{c}_{s-1}, \hat{c}_{s-2}, \ldots, \hat{c}_1\}$.
We then compute the mean IoU between the reverse track and the prompt object track over all frames prior to $s$:
\begin{equation}
    S_\text{rev}(C, P) = \frac{1}{s - 1} \sum_{t=1}^{s-1} \frac{|\hat{c}_t \cap p_t|}{|\hat{c}_t \cup p_t|}.
    \label{eq:rev}
\end{equation}
The intuition is straightforward: if a candidate region is a genuine transformation product of the prompt object, propagating it backward should align with the original prompt object.
This is precisely the behavior observed for Candidate~\#1 in Figure~\ref{fig:pipeline}(a), whose reverse track lands on the foil box and yields $S_\text{rev} = 0.7617$; in contrast, Candidate~\#2's reverse track stays on the actor's hands, yielding $S_\text{rev} = 0.0002$.
We therefore retain only candidates for which $S_\text{rev}(C, P) > \tau_\text{rev}$, and discard the rest.
The final tracking result $\mathcal{T}$ is formed by combining the surviving candidate tracks with the original prompt track $P$ with VLM prompting, exactly as in \prev --- visualized as the merged tubes in Figure~\ref{fig:pipeline}(b).

\subsection{\prev vs.~\ours}
\label{sec:method-summary}

Table~\ref{tab:v1v2} summarizes the key differences between \prev and \ours.
The central shift is from an \emph{eager} strategy --- partitioning the entire video upfront --- to a \emph{reactive} one that queries entity segmentation and initiates new tracks only when SAM2's internal hypothesis suggests its necessity.
This eliminates the two primary bottlenecks of \prev: the majority of the per-frame entity segmentation is avoided by a lightweight heuristic check, and the number of tracked objects at any time is bounded by the number of detected candidates rather than the total number of scene entities.
Furthermore, the semantic consistency check requiring FC-CLIP is replaced entirely by reverse tracking, which reuses SAM2 and introduces no additional model dependency.
The state graph construction via VLM querying remains unchanged from \prev.

\begin{table}[ht]
\centering

\resizebox{0.47\textwidth}{!}{%
\begin{tabular}{lll}
\toprule
\textbf{Aspect} & \textbf{\prev} & \textbf{\ours} \\
\midrule
Entity queries  & Every frame              & Selected frames only \\
Tracking scaling    & \# entities in video         & \# detected candidates \\
Candidate discovery & Object emergence         & Multi-mask disagreement \\
Candidate filtering & Proximity + FC-CLIP      & Reverse tracking \\
\bottomrule
\end{tabular}}
\caption{Comparison of \prev and \ours.}
\label{tab:v1v2}
\end{table}
\begin{table*}[ht]
\centering
\resizebox{\textwidth}{!}{%
\begin{tabular}{ll c cc @{\hspace{9pt}}ccccc  c}
\toprule
\multirow{2}{*}{\textbf{Dataset}} & \multirow{2}{*}{\textbf{Method}} & \textbf{Detect} & \multirow{2}{*}{$\mathcal{J}$} & \multirow{2}{*}{$\mathcal{J}$\textsubscript{tr}} & & \multicolumn{4}{c}{\textbf{Time per comp-mask (s)}} & \multirow{2}{*}{\textbf{MPS}} \\
\cmidrule(lr){7-10}
 &  & \textbf{Transform.} & & & & Entity Seg. & Track & FC-CLIP & Total & \\
\midrule
\multirow{3}{*}{VOST val}
  & SAM2.1~\cite{sam2}        &  &  48.4          & 32.4 & & \gray --     & \gray 0.283  & \gray --     & \gray 0.283           & \gray 3.53 \\
  & \prev~\cite{tubeletgraph} & \cmark & 50.9          & 36.7 & & 1.097  & 3.228  & 0.067  & 4.392           & 0.23 \\
  & \ours                     & \cmark & \textbf{52.8} & \textbf{36.8} & & 0.300  & 1.031  & --     & \textbf{1.331}  & \textbf{0.75} \\
\midrule
\multirow{3}{*}{VSCOS val}
  & SAM2.1~\cite{sam2}        &  &  72.0          &  66.9 & & \gray --     & \gray 0.264  & \gray --     & \gray 0.264           & \gray 3.79 \\
  & \prev~\cite{tubeletgraph} & \cmark & \textbf{75.9} & \textbf{72.2} & & 1.030  & 2.551  & 0.065  & 3.646           & 0.27 \\
  & \ours                     & \cmark & 75.3          & 70.7 & & 0.109  & 0.686  & --     & \textbf{0.795}  & \textbf{1.26} \\
\midrule
\multirow{3}{*}{M\textsuperscript{3}-VOS val}
  & SAM2.1~\cite{sam2}        &  &  71.3          &  59.3 & & \gray --     & \gray 0.280  & \gray --     & \gray 0.280           & \gray 3.57 \\
  & \prev~\cite{tubeletgraph} & \cmark & \textbf{74.1} & \textbf{64.1} & & 1.166  & 1.997  & 0.062  & 3.225           & 0.31 \\
  & \ours                     & \cmark & 73.7          & 63.8 & & 0.129  & 0.740  & --     & \textbf{0.869}  & \textbf{1.15} \\
\midrule
\multirow{3}{*}{DAVIS17 val}
  & SAM2.1~\cite{sam2}        &  &  85.7          &  83.0 & & \gray --     & \gray 0.304  & \gray --     & \gray 0.304           & \gray 3.29 \\
  & \prev~\cite{tubeletgraph} & \cmark & 85.6          & 82.6 & & 1.213  & 3.558  & 0.034  & 4.805           & 0.21 \\
  & \ours                     & \cmark & \textbf{85.8} & \textbf{83.2} & & 0.071  & 0.379  & --     & \textbf{0.450}  & \textbf{2.22} \\
\bottomrule
\end{tabular}
}
\caption{Per-component inference cost. Time per comp-mask is reported for entity segmentation, tracking, and semantic filtering via FC-CLIP~\cite{fcclip}. MPS reflects total masks produced per second across all tracked objects. SAM2~\cite{sam2} timings are in gray as reference.}
\label{tab:efficiency}
\end{table*}

\section{Experiments}
\label{sec:exp}

\subsection{Experimental Setup}
\label{sec:exp-setup}

\textbf{Datasets.}
Following \prev~\cite{tubeletgraph}, we evaluate on four object tracking benchmarks.
VOST~\cite{vost} validation set contains 70 ego-centric videos from EPIC-Kitchens~\cite{ek100} and Ego4D~\cite{ego4d} focusing on objects under transformations.
Similarly, VSCOS~\cite{vscos} val contains 98 ego-centric videos focused on state-changing object tracking.
M\textsuperscript{3}-VOS~\cite{m3vos} val contains 479 videos depicting object phase transitions.
In addition, we evaluate on DAVIS17~\cite{davis17} as a sanity check for transformation-free videos.
We evaluate state graph on VOST-TAS~\cite{tubeletgraph}, an extension of the VOST validation set with 57 video instances and 108 transformations.

\textbf{Metrics.}
For tracking, we follow VOST~\cite{vost} and report Jaccard $\mathcal{J}$ over the entire video and $\mathcal{J}_\text{tr}$ over the last 25\% of frames (where transformations have typically completed).
For state graph evaluation, we follow \prev~\cite{tubeletgraph} and report semantic accuracy of predicted action verbs ($S_V$) and resulting object descriptions ($S_O$), temporal localization precision and recall ($T_P$, $T_R$), spatiotemporal recall $H_\text{ST}$ (correct temporal match with all resulting objects matched at IoU $> 0.5$), and overall recall $H$ (additionally requiring correct semantic descriptions).
To evaluate efficiency, we use the total number of object component masks (comp-masks) predicted as normalization, since the system may need to keep track of multiple objects that emerge at various points in time.
We report wall-clock time per comp-mask broken down to separate modules where applicable, along with total mask per second (MPS).
All timings are measured on a single NVIDIA RTX A6000 GPU.

\textbf{Implementation Details.}
\ours shares its backbone models with \prev: SAM2.1-L~\cite{sam2} for tracking, CropFormer-Hornet3X~\cite{cropformer} for entity segmentation, and GPT-4.1~\cite{gpt4} (sampling temperature 0) for state graph construction.
We select $\tau_\text{agree} = 0.75$, $\tau_\text{res} = 0.1$, $\tau_\text{new} = 0.3$ and $\tau_\text{rev} = 0.1$ by sweeping on VOST training split, and apply to all other datasets without further tuning.

\begin{table*}[t]
\centering
\resizebox{\textwidth}{!}{%
\begin{tabular}{lc c@{\hspace{6pt}}c@{\hspace{6pt}}c@{\hspace{6pt}}c@{\hspace{6pt}}c@{\hspace{6pt}}c c@{\hspace{6pt}}c@{\hspace{6pt}}c@{\hspace{6pt}}c@{\hspace{6pt}}c@{\hspace{6pt}}c}
\toprule
 & &\multicolumn{6}{c}{\textbf{Tracking}} &\multicolumn{6}{c}{\textbf{State Graph}}\\
 \pad \cline{3-14} \pad
\textbf{Method} & \textbf{Detect} &\multicolumn{2}{c}{VOST} &\multicolumn{2}{c}{VSCOS} &\multicolumn{2}{c}{M$^3$-VOS} &\multicolumn{2}{c}{Sem. Acc.} &\multicolumn{2}{c}{Temp. Loc.} &\multicolumn{2}{c}{Overall}\\
\pad \cline{3-14} \pad
 & \textbf{Transf.} & $\mathcal{J}$ & $\mathcal{J}$\textsubscript{tr} & $\mathcal{J}$ & $\mathcal{J}$\textsubscript{tr} & $\mathcal{J}$ & $\mathcal{J}$\textsubscript{tr} & $S_V$ & $S_O$ & $T_P$ & $T_R$ & $H_{ST}$ & $H$ \\
\midrule
Cutie~\cite{cutie}  &          &41.1	&25.5	&70.9	&67.1	&74.5	&64.7	& - & - & - & - & - & - \\
ReVOS~\cite{m3vos} &           &41.0	&25.3	&-	    &-	    &\textbf{75.6}	&\textbf{66.5}	& - & - & - & - & - & - \\
SAM2~\cite{sam2} &            &46.1	&29.4	&72.5	&67.1	&71.3	&59.8	& - & - & - & - & - & - \\
SAM2Long~\cite{sam2long} &        &46.4	&29.1	&73.0	&68.6	&70.2	&58.7	& - & - & - & - & - & - \\
SAM2.1~\cite{sam2}	&         &48.4	&32.4	&72.0	&66.9	&71.3	&59.3	& - & - & - & - & - & - \\
DAM4SAM~\cite{dam4sam} &  &48.8	&33.6	&71.3	&66.0	&72.2	&61.3	& - & - & - & - & - & - \\
SAMURAI~\cite{samurai} &         &49.8	&34.0	&71.8	&66.9	&72.6	&61.6	& - & - & - & - & - & - \\
\midrule
\prev~\cite{tubeletgraph} & \cmark & 50.9   & 36.7 & \textbf{75.9} & \textbf{72.2} & 74.1 & 64.1  &\gray \textbf{81.8} &\gray 72.3 &\textbf{43.1} &\textbf{20.4}  &\textbf{12.0} &6.5\\
\ours                     & \cmark & \textbf{52.8} & \textbf{36.8} & 75.3 & 70.7 & 73.7 & 63.8 &\gray 77.3 &\gray \textbf{82.9} & 36.7 & \textbf{20.4} & \textbf{12.0} &\textbf{9.3} \\
\bottomrule
\end{tabular}
}
\caption{Quantitative results for object tracking on VOST~\cite{vost}, VSCOS~\cite{vscos} and M\textsuperscript{3}-VOS~\cite{m3vos} and state graph on VOST-TAS~\cite{tubeletgraph}.}
\label{tab:benchmark}
\end{table*}

\subsection{Computational Efficiency}
\label{sec:exp-efficiency}

We first jointly evaluate \ours's computational efficiency and tracking quality against \prev~\cite{tubeletgraph} and the base SAM2.1~\cite{sam2} tracker.
Table~\ref{tab:efficiency} reports per-component inference cost, overall mask per second (MPS), and tracking performance across all four benchmarks.

\textbf{\ours achieves consistent speedups.}
On VOST, \ours reduces inference cost from 4.39 seconds per comp-mask to 1.33, bringing the MPS from 0.23 to 0.75  ($\sim$3.3$\times$ speedup).
Similar time reductions are also observed on VSCOS (3.65 to 0.80, $\sim$4.6$\times$), M\textsuperscript{3}-VOS (3.23 to 0.87, $\sim$3.7$\times$), and DAVIS17 (4.81 to 0.45, $\sim$10.7$\times$).
In terms of per-component inference costs, the entity segmentation cost collapses dramatically and sees an average $\sim$9.8$\times$ reduction across four benchmarks.
At the same time, the tracking cost shows a similar trend, dropping by $\sim$3$\times$ on VOST and nearly $10\times$ on DAVIS17.
These reductions confirm the superiority of the \ours's reactive design, where unnecessary cost of entity segmentation and object tracking is significantly reduced.

\textbf{Efficiency gains come at no trade-off to tracking quality.}
On VOST, \ours achieves $\mathcal{J} = 52.8$ versus 50.9 for \prev --- a $+1.9$ improvement despite running $\sim$3.3$\times$ faster.
On VSCOS and M\textsuperscript{3}-VOS, \ours is within $0.6$ $\mathcal{J}$ of \prev with no statistically meaningful gap.
We attribute the VOST gain to the reactive nature of candidate discovery: by focusing only on regions flagged by SAM2's own uncertainty, \ours introduces fewer false positives than \prev's exhaustive partition.

\textbf{Graceful behavior on transformation-free videos.}
The DAVIS17 result is particularly informative as a sanity check.
Since DAVIS17 contains no object transformations, the multi-mask disagreement signal rarely fires and almost no CropFormer queries are made.
As a result, \ours costs only 0.45 seconds per component mask (a mere $1.5\times$ overhead over SAM2.1) --- compared to a $16\times$ cost for \prev at 4.81 seconds.
Moreover, tracking quality is preserved on DAVIS, confirming that \ours pays for transformation handling only when transformations actually occur, without inducing any quality regression on transformation-free videos.

\subsection{Tracking and State-Graph Performance}
\label{sec:exp-sota}
We evaluate FluxGraph against a set of trackers and report tracking results across three transformation benchmarks (VOST, VSCOS, M\textsuperscript{3}-VOS), as well as evaluate state graph construction on VOST-TAS in Table~\ref{tab:benchmark} .

\textbf{Tracking under transformations.}
\ours outperforms all trackers on VOST ($\mathcal{J} = 52.8$, $+3.0$ over SAMURAI) while uniquely providing state graph predictions.
In addition, \ours demonstrates competitive performance compared to \prev, with \ours within $0.6$ $\mathcal{J}$ of \prev and $\sim$3--5$\times$ faster.

\begin{table}[t]
\centering
\resizebox{0.478\textwidth}{!}{%
\begin{tabular}{l @{\hspace{5pt}}c@{\hspace{5pt}}c@{\hspace{7pt}} c@{\hspace{5pt}}c c cc}
\toprule
& Reactive & Rev. & \multicolumn{2}{c}{\textbf{Time / mask (s)}} & \multirow{2}{*}{\textbf{MPS}} & \multirow{2}{*}{$\mathcal{J}$} & \multirow{2}{*}{$\mathcal{J}$\textsubscript{tr}} \\
\cmidrule(lr){4-5}
 & Discovery & Track & Entity Seg. & Track & & & \\
\midrule
\multirow{3}{*}{\rotatebox[origin=c]{90}{VOST}}
  & \cmark &        & --    & --    & --            & 37.8          & 25.4 \\
  &        & \cmark & 1.014 & 1.068 & 0.48          & 53.1          & 37.3 \\
  & \cmark & \cmark & 0.300 & 1.031 & 0.75          & 52.8          & 36.8 \\
\midrule
\multirow{3}{*}{\rotatebox[origin=c]{90}{VSCOS}}
  & \cmark &        & --    & --    & --            & 72.0          & 65.9 \\
  &        & \cmark & 0.981 & 1.004 & 0.50          & 75.2          & 70.4 \\
  & \cmark & \cmark & 0.109 & 0.686 & 1.26          & 75.3          & 70.7 \\
\midrule
\multirow{3}{*}{\rotatebox[origin=c]{90}{M\textsuperscript{3}-VOS}}
  & \cmark &        & --    & --    & --            & 67.8          & 57.7 \\
  &        & \cmark & 1.152 & 0.999 & 0.46          & 73.4          & 63.4 \\
  & \cmark & \cmark & 0.129 & 0.740 & 1.15          & 73.7          & 63.8 \\
\midrule
\multirow{3}{*}{\rotatebox[origin=c]{90}{DAVIS}}
  & \cmark &        & --    & --    & --            & 80.3          & 72.7 \\
  &        & \cmark & 1.195 & 0.388 & 0.63          & 85.6          & 82.7 \\
  & \cmark & \cmark & 0.071 & 0.379 & 2.22          & 85.8          & 83.2 \\
\bottomrule
\end{tabular}
}
\caption{Ablation study on \ours components.\vspace{-0.5cm}}
\label{tab:ablation}
\end{table}

\textbf{State graph performance.}
\ours matches \prev's spatiotemporal recall ($H_\text{ST} = 12.0$) and temporal recall ($T_R = 20.4$), effectively preserving the set of correctly-localized transformations.
Notably, \ours achieves a higher overall recall ($H = 9.3$ vs.~6.5 for \prev), and substantially improves the semantic accuracy of resulting object descriptions ($S_O = 82.9$ vs.~72.3 for \prev).
We attribute the improved object-description accuracy to \ours's tighter candidate proposals: because the reactive trigger restricts entity segmentation to multi-mask disagreement regions, the resulting object masks are more clearly localized, providing a sharper visual prompt for the VLM.

\begin{figure*}[t]
  \centering
  \includegraphics[width=\linewidth]{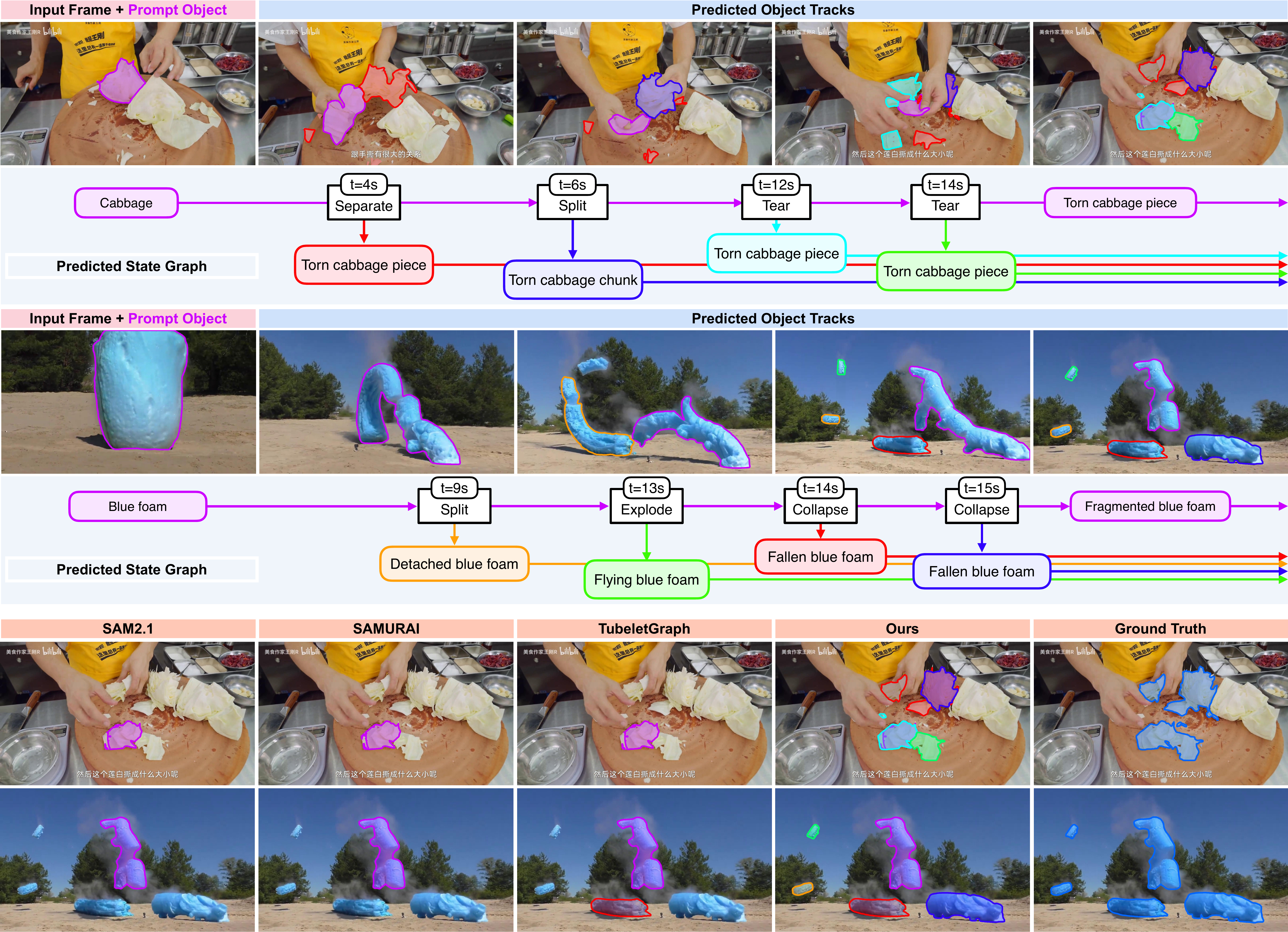}
  \caption{%
    Qualitative results. We showcase \ours{}’s tracking and state graph predictions on top, with comparisons against baselines at an ending frame at the bottom.
  }
  \label{fig:qual}
  \vspace{-0.25cm}
\end{figure*}

\subsection{Ablation Study}
\label{sec:exp-ablation}

We ablate the two core components of \ours --- the reactive candidate discovery heuristic (Sec.~\ref{sec:method-discovery}) and reverse tracking (Sec.~\ref{sec:method-filtering}) --- by separately removing each.
Results across all four benchmarks are reported in Table~\ref{tab:ablation}.

\textbf{Reverse tracking is essential for tracking quality.}
Without reverse tracking, \ours admits any candidate proposed by the multi-mask trigger, resulting in a substantial quality collapse: $\mathcal{J}$ drops from 52.8 to 37.8 on VOST, 75.3 to 72.0 on VSCOS, 73.7 to 67.8 on M\textsuperscript{3}-VOS , and 85.8 to 80.3 on DAVIS17.
Reverse tracking is therefore necessary in maintaining tracking quality, effectively replacing the proximity and semantic priors of \prev.

\textbf{Reactive discovery is a strong efficiency--quality tradeoff.}
Comparing \ours with and without reactive candidate discovery, we see that skipping all down-stream analysis when possible reduces total inference time substantially: on VOST, MPS improves from 0.48 to 0.75; on VSCOS from 0.50 to 1.26; on M\textsuperscript{3}-VOS from 0.46 to 1.15; and on DAVIS17 from 0.63 to 2.22.
Across all four datasets, tracking quality changes by less than $0.3$ $\mathcal{J}$ when it is enabled.
This heuristic therefore delivers the bulk of \ours's efficiency gain at negligible quality cost.

\textbf{Where the residual cost goes.}
Even with both components enabled, the tracking term remains the largest single contributor to \ours's cost (\eg 1.03 s/comp-mask on VOST out of 1.33 total).
This is attributed to the cost of tracking false-positive candidates, which is necessary for reasoning about their removal from the final prediction.

\subsection{Qualitative Results}
\label{sec:exp-qualitative}

Figure~\ref{fig:qual} showcases two challenging transformation scenarios: a cabbage torn into pieces, and an expansive blue foam splits and collapses mid-air.
On both, \ours reliably detects each transformation and produces a clean track for every emerging component.
In contrast, SAM2.1~\cite{sam2} and SAMURAI~\cite{samurai} fail to recover any of the resulting components, while TubeletGraph~\cite{tubeletgraph} recovers only a limited fraction.
These results further confirm the efficacy of multi-mask disagreement as a lightweight transformation trigger.
\section{Conclusion}
\label{sec:conclusion}

In this work, we proposed \ours, an efficient, reactive variant of \prev~\cite{tubeletgraph} for tracking objects through transformations and constructing state graphs describing each observed change.
Building on the observation that SAM2's internal multi-mask outputs already encode where related objects are likely to emerge, \ours replaces \prev's eager spatiotemporal partition with a reactive trigger based on multi-mask disagreement, and replaces the proximity and semantic check with model-free reverse tracking.
Across four benchmarks, \ours achieves 3.3--10.7$\times$ speedups over \prev while matching tracking and state graph quality.

\textbf{Limitations and Broader Impacts.}
Beyond the limitations inherited from \prev (\eg only first-order transformations can be detected), \ours introduces two additional caveats.
First, fast transformations that produce minimal multi-mask disagreement may evade the reactive trigger, leading to missed detections that \prev's exhaustive partition would catch.
Second, \ours's core insight depends on SAM2's alternative heads that produce meaningful hypotheses; future trackers that do not expose such internal uncertainty would require a different trigger signal.
Finally, deployment in surveillance or human-monitoring contexts should carefully consider the rights of bystanders, particularly in the ego-centric settings where these benchmarks are sourced.

% \textbf{Acknowledgment.} 
% %
% This research is based upon work supported in part by the National Science Foundation (IIS-2144117 and IIS-2107161).
% Yihong Sun is supported by an NSF graduate research fellowship.
{
    \small
    \bibliographystyle{ieeenat_fullname}
    \bibliography{main}

@String(ECCV= {Eur. Conf. Comput. Vis.})

@String(ECCV  = {ECCV})

@article{m3vos,
  title={M$^3$-VOS: Multi-Phase, Multi-Transition, and Multi-Scenery Video Object Segmentation},
  author={Chen, Zixuan and Li, Jiaxin and Tan, Liming and Guo, Yejie and Liang, Junxuan and Lu, Cewu and Li, Yong-Lu},
  journal={arXiv preprint arXiv:2412.13803},
  year={2024}
}

@inproceedings{vscos,
  title={Video state-changing object segmentation},
  author={Yu, Jiangwei and Li, Xiang and Zhao, Xinran and Zhang, Hongming and Wang, Yu-Xiong},
  booktitle={Proceedings of the IEEE/CVF International Conference on Computer Vision},
  pages={20439--20448},
  year={2023}
}

@inproceedings{vost,
  title={Breaking the" object" in video object segmentation},
  author={Tokmakov, Pavel and Li, Jie and Gaidon, Adrien},
  booktitle={Proceedings of the IEEE/CVF Conference on Computer Vision and Pattern Recognition},
  pages={22836--22845},
  year={2023}
}

@article{sam2,
  title={Sam 2: Segment anything in images and videos},
  author={Ravi, Nikhila and Gabeur, Valentin and Hu, Yuan-Ting and Hu, Ronghang and Ryali, Chaitanya and Ma, Tengyu and Khedr, Haitham and R{\"a}dle, Roman and Rolland, Chloe and Gustafson, Laura and others},
  journal={arXiv preprint arXiv:2408.00714},
  year={2024}
}

@misc{oscs,
      title={Learning Object State Changes in Videos: An Open-World Perspective}, 
      author={Zihui Xue and Kumar Ashutosh and Kristen Grauman},
      year={2024},
      eprint={2312.11782},
      archivePrefix={arXiv},
      primaryClass={cs.CV},
      url={https://arxiv.org/abs/2312.11782}, 
}

@misc{spoc,
      title={SPOC: Spatially-Progressing Object State Change Segmentation in Video}, 
      author={Priyanka Mandikal and Tushar Nagarajan and Alex Stoken and Zihui Xue and Kristen Grauman},
      year={2025},
      eprint={2503.11953},
      archivePrefix={arXiv},
      primaryClass={cs.CV},
      url={https://arxiv.org/abs/2503.11953}, 
}

@inproceedings{liu2024blade,
  title={BLADE: Learning Compositional Behaviors from Demonstration and Language},
  author={Liu, Weiyu and Nie, Neil and Zhang, Ruohan and Mao, Jiayuan and Wu, Jiajun},
  booktitle={Conference on Robot Learning (CoRL)},
  year={2024}
}

@article{davis17,
  title={The 2017 davis challenge on video object segmentation},
  author={Pont-Tuset, Jordi and Perazzi, Federico and Caelles, Sergi and Arbel{\'a}ez, Pablo and Sorkine-Hornung, Alex and Van Gool, Luc},
  journal={arXiv preprint arXiv:1704.00675},
  year={2017}
}

@inproceedings{segmentanything,
  title={Segment anything},
  author={Kirillov, Alexander and Mintun, Eric and Ravi, Nikhila and Mao, Hanzi and Rolland, Chloe and Gustafson, Laura and Xiao, Tete and Whitehead, Spencer and Berg, Alexander C and Lo, Wan-Yen and others},
  booktitle={Proceedings of the IEEE/CVF international conference on computer vision},
  pages={4015--4026},
  year={2023}
}

@article{cropformer,
  title={High-quality entity segmentation},
  author={Qi, Lu and Kuen, Jason and Guo, Weidong and Shen, Tiancheng and Gu, Jiuxiang and Jia, Jiaya and Lin, Zhe and Yang, Ming-Hsuan},
  journal={arXiv preprint arXiv:2211.05776},
  year={2022}
}

@article{fcclip,
  title={Convolutions die hard: Open-vocabulary segmentation with single frozen convolutional clip},
  author={Yu, Qihang and He, Ju and Deng, Xueqing and Shen, Xiaohui and Chen, Liang-Chieh},
  journal={Advances in Neural Information Processing Systems},
  volume={36},
  pages={32215--32234},
  year={2023}
}

@article{gpt4,
  title={Gpt-4 technical report},
  author={Achiam, Josh and Adler, Steven and Agarwal, Sandhini and Ahmad, Lama and Akkaya, Ilge and Aleman, Florencia Leoni and Almeida, Diogo and Altenschmidt, Janko and Altman, Sam and Anadkat, Shyamal and others},
  journal={arXiv preprint arXiv:2303.08774},
  year={2023}
}

@inproceedings{vot2016,
  title={The visual object tracking VOT2016 challenge results},
  author={Roffo, Giorgio and Melzi, Simone and others},
  booktitle={Computer Vision--ECCV 2016 Workshops: Amsterdam, The Netherlands, October 8-10 and 15-16, 2016, Proceedings, Part II},
  pages={777--823},
  year={2016},
  organization={Springer International Publishing}
}

@inproceedings{davis16,
  title={A benchmark dataset and evaluation methodology for video object segmentation},
  author={Perazzi, Federico and Pont-Tuset, Jordi and McWilliams, Brian and Van Gool, Luc and Gross, Markus and Sorkine-Hornung, Alexander},
  booktitle={Proceedings of the IEEE conference on computer vision and pattern recognition},
  pages={724--732},
  year={2016}
}

@inproceedings{ytvos,
  title={Youtube-vos: Sequence-to-sequence video object segmentation},
  author={Xu, Ning and Yang, Linjie and Fan, Yuchen and Yang, Jianchao and Yue, Dingcheng and Liang, Yuchen and Price, Brian and Cohen, Scott and Huang, Thomas},
  booktitle={Proceedings of the European conference on computer vision (ECCV)},
  pages={585--601},
  year={2018}
}

@inproceedings{lvos,
  title={Lvos: A benchmark for long-term video object segmentation},
  author={Hong, Lingyi and Chen, Wenchao and Liu, Zhongying and Zhang, Wei and Guo, Pinxue and Chen, Zhaoyu and Zhang, Wenqiang},
  booktitle={Proceedings of the IEEE/CVF International Conference on Computer Vision},
  pages={13480--13492},
  year={2023}
}

@inproceedings{mose,
  title={MOSE: A new dataset for video object segmentation in complex scenes},
  author={Ding, Henghui and Liu, Chang and He, Shuting and Jiang, Xudong and Torr, Philip HS and Bai, Song},
  booktitle={Proceedings of the IEEE/CVF international conference on computer vision},
  pages={20224--20234},
  year={2023}
}

@inproceedings{mevis,
  title={MeViS: A large-scale benchmark for video segmentation with motion expressions},
  author={Ding, Henghui and Liu, Chang and He, Shuting and Jiang, Xudong and Loy, Chen Change},
  booktitle={Proceedings of the IEEE/CVF international conference on computer vision},
  pages={2694--2703},
  year={2023}
}

@inproceedings{lasot,
  title={Lasot: A high-quality benchmark for large-scale single object tracking},
  author={Fan, Heng and Lin, Liting and Yang, Fan and Chu, Peng and Deng, Ge and Yu, Sijia and Bai, Hexin and Xu, Yong and Liao, Chunyuan and Ling, Haibin},
  booktitle={Proceedings of the IEEE/CVF conference on computer vision and pattern recognition},
  pages={5374--5383},
  year={2019}
}

@inproceedings{bhat2020learning,
  title={Learning what to learn for video object segmentation},
  author={Bhat, Goutam and Lawin, Felix J{\"a}remo and Danelljan, Martin and Robinson, Andreas and Felsberg, Michael and Van Gool, Luc and Timofte, Radu},
  booktitle={Computer Vision--ECCV 2020: 16th European Conference, Glasgow, UK, August 23--28, 2020, Proceedings, Part II 16},
  pages={777--794},
  year={2020},
  organization={Springer}
}

@inproceedings{caelles2017one,
  title={One-shot video object segmentation},
  author={Caelles, Sergi and Maninis, Kevis-Kokitsi and Pont-Tuset, Jordi and Leal-Taix{\'e}, Laura and Cremers, Daniel and Van Gool, Luc},
  booktitle={Proceedings of the IEEE conference on computer vision and pattern recognition},
  pages={221--230},
  year={2017}
}

@article{maninis2018video,
  title={Video object segmentation without temporal information},
  author={Maninis, K-K and Caelles, Sergi and Chen, Yuhua and Pont-Tuset, Jordi and Leal-Taix{\'e}, Laura and Cremers, Daniel and Van Gool, Luc},
  journal={IEEE transactions on pattern analysis and machine intelligence},
  volume={41},
  number={6},
  pages={1515--1530},
  year={2018},
  publisher={IEEE}
}

@inproceedings{hu2018videomatch,
  title={Videomatch: Matching based video object segmentation},
  author={Hu, Yuan-Ting and Huang, Jia-Bin and Schwing, Alexander G},
  booktitle={Proceedings of the European conference on computer vision (ECCV)},
  pages={54--70},
  year={2018}
}

@inproceedings{voigtlaender2019feelvos,
  title={Feelvos: Fast end-to-end embedding learning for video object segmentation},
  author={Voigtlaender, Paul and Chai, Yuning and Schroff, Florian and Adam, Hartwig and Leibe, Bastian and Chen, Liang-Chieh},
  booktitle={Proceedings of the IEEE/CVF conference on computer vision and pattern recognition},
  pages={9481--9490},
  year={2019}
}

@inproceedings{yang2018efficient,
  title={Efficient video object segmentation via network modulation},
  author={Yang, Linjie and Wang, Yanran and Xiong, Xuehan and Yang, Jianchao and Katsaggelos, Aggelos K},
  booktitle={Proceedings of the IEEE conference on computer vision and pattern recognition},
  pages={6499--6507},
  year={2018}
}

@inproceedings{xmem,
  title={Xmem: Long-term video object segmentation with an atkinson-shiffrin memory model},
  author={Cheng, Ho Kei and Schwing, Alexander G},
  booktitle={European Conference on Computer Vision},
  pages={640--658},
  year={2022},
  organization={Springer}
}

@inproceedings{oh2018fast,
  title={Fast video object segmentation by reference-guided mask propagation},
  author={Oh, Seoung Wug and Lee, Joon-Young and Sunkavalli, Kalyan and Kim, Seon Joo},
  booktitle={Proceedings of the IEEE conference on computer vision and pattern recognition},
  pages={7376--7385},
  year={2018}
}

@article{yang2021associating,
  title={Associating objects with transformers for video object segmentation},
  author={Yang, Zongxin and Wei, Yunchao and Yang, Yi},
  journal={Advances in Neural Information Processing Systems},
  volume={34},
  pages={2491--2502},
  year={2021}
}

@inproceedings{cutie,
  title={Putting the object back into video object segmentation},
  author={Cheng, Ho Kei and Oh, Seoung Wug and Price, Brian and Lee, Joon-Young and Schwing, Alexander},
  booktitle={Proceedings of the IEEE/CVF Conference on Computer Vision and Pattern Recognition},
  pages={3151--3161},
  year={2024}
}

@inproceedings{oh2019video,
  title={Video object segmentation using space-time memory networks},
  author={Oh, Seoung Wug and Lee, Joon-Young and Xu, Ning and Kim, Seon Joo},
  booktitle={Proceedings of the IEEE/CVF international conference on computer vision},
  pages={9226--9235},
  year={2019}
}

@inproceedings{yang2020collaborative,
  title={Collaborative video object segmentation by foreground-background integration},
  author={Yang, Zongxin and Wei, Yunchao and Yang, Yi},
  booktitle={European Conference on Computer Vision},
  pages={332--348},
  year={2020},
  organization={Springer}
}

@article{sam2long,
  title={Sam2long: Enhancing sam 2 for long video segmentation with a training-free memory tree},
  author={Ding, Shuangrui and Qian, Rui and Dong, Xiaoyi and Zhang, Pan and Zang, Yuhang and Cao, Yuhang and Guo, Yuwei and Lin, Dahua and Wang, Jiaqi},
  journal={arXiv preprint arXiv:2410.16268},
  year={2024}
}

@article{samurai,
  title={Samurai: Adapting segment anything model for zero-shot visual tracking with motion-aware memory},
  author={Yang, Cheng-Yen and Huang, Hsiang-Wei and Chai, Wenhao and Jiang, Zhongyu and Hwang, Jenq-Neng},
  journal={arXiv preprint arXiv:2411.11922},
  year={2024}
}

@article{ek100,
  title={Rescaling egocentric vision: Collection, pipeline and challenges for epic-kitchens-100},
  author={Damen, Dima and Doughty, Hazel and Farinella, Giovanni Maria and Furnari, Antonino and Kazakos, Evangelos and Ma, Jian and Moltisanti, Davide and Munro, Jonathan and Perrett, Toby and Price, Will and others},
  journal={International Journal of Computer Vision},
  pages={1--23},
  year={2022},
  publisher={Springer}
}

@inproceedings{ego4d,
  title={Ego4d: Around the world in 3,000 hours of egocentric video},
  author={Grauman, Kristen and Westbury, Andrew and Byrne, Eugene and Chavis, Zachary and Furnari, Antonino and Girdhar, Rohit and Hamburger, Jackson and Jiang, Hao and Liu, Miao and Liu, Xingyu and others},
  booktitle={Proceedings of the IEEE/CVF conference on computer vision and pattern recognition},
  pages={18995--19012},
  year={2022}
}

@inproceedings{dam4sam,
  title={A distractor-aware memory for visual object tracking with sam2},
  author={Videnovic, Jovana and Lukezic, Alan and Kristan, Matej},
  booktitle={Proceedings of the Computer Vision and Pattern Recognition Conference},
  pages={24255--24264},
  year={2025}
}

@article{sparta,
  title={Mash, Spread, Slice! Learning to Manipulate Object States via Visual Spatial Progress},
  author={Mandikal, Priyanka and Hu, Jiaheng and Dass, Shivin and Majumder, Sagnik and Mart{\'\i}n-Mart{\'\i}n, Roberto and Grauman, Kristen},
  journal={arXiv preprint arXiv:2509.24129},
  year={2025}
}

@inproceedings{wu2024tracking,
  title={Tracking transforming objects: A benchmark},
  author={Wu, You and Wang, Yuelong and Liao, Yaxin and Wu, Fuliang and Ye, Hengzhou and Li, Shuiwang},
  booktitle={Chinese Conference on Pattern Recognition and Computer Vision (PRCV)},
  pages={222--236},
  year={2024},
  organization={Springer}
}

@article{tubeletgraph,
  title={Tracking and Understanding Object Transformations},
  author={Sun, Yihong and Yang, Xinyu and Sun, Jennifer J and Hariharan, Bharath},
  journal={arXiv preprint arXiv:2511.04678},
  year={2025}
}

@article{trajtok,
  title={TrajTok: Learning Trajectory Tokens enables better Video Understanding},
  author={Zheng, Chenhao and Zhang, Jieyu and Zhang, Jianing and Huang, Weikai and Kumar, Ashutosh and Kong, Quan and Tuzel, Oncel and Li, Chun-Liang and Krishna, Ranjay},
  journal={arXiv preprint arXiv:2602.22779},
  year={2026}
}

@article{wildlife,
  title={A literature review of computer vision techniques in wildlife monitoring},
  author={Neupane, Sangam B and Sato, Kazuhiko and Gautam, Bishnu P},
  journal={IJSRP},
  volume={16},
  pages={282--295},
  year={2022}
}

@inproceedings{habibian2021skip,
  title={Skip-convolutions for efficient video processing},
  author={Habibian, Amirhossein and Abati, Davide and Cohen, Taco S and Bejnordi, Babak Ehteshami},
  booktitle={Proceedings of the IEEE/CVF Conference on computer vision and pattern recognition},
  pages={2695--2704},
  year={2021}
}

@inproceedings{dutson2022event,
  title={Event neural networks},
  author={Dutson, Matthew and Li, Yin and Gupta, Mohit},
  booktitle={European Conference on Computer Vision},
  pages={276--293},
  year={2022},
  organization={Springer}
}

@inproceedings{habibian2022delta,
  title={Delta distillation for efficient video processing},
  author={Habibian, Amirhossein and Ben Yahia, Haitam and Abati, Davide and Gavves, Efstratios and Porikli, Fatih},
  booktitle={European Conference on Computer Vision},
  pages={213--229},
  year={2022},
  organization={Springer}
}

@inproceedings{wu2019adaframe,
  title={Adaframe: Adaptive frame selection for fast video recognition},
  author={Wu, Zuxuan and Xiong, Caiming and Ma, Chih-Yao and Socher, Richard and Davis, Larry S},
  booktitle={Proceedings of the IEEE/CVF Conference on Computer Vision and Pattern Recognition},
  pages={1278--1287},
  year={2019}
}

@inproceedings{korbar2019scsampler,
  title={Scsampler: Sampling salient clips from video for efficient action recognition},
  author={Korbar, Bruno and Tran, Du and Torresani, Lorenzo},
  booktitle={Proceedings of the IEEE/CVF International Conference on Computer Vision},
  pages={6232--6242},
  year={2019}
}

@inproceedings{wang2021adaptive,
  title={Adaptive focus for efficient video recognition},
  author={Wang, Yulin and Chen, Zhaoxi and Jiang, Haojun and Song, Shiji and Han, Yizeng and Huang, Gao},
  booktitle={proceedings of the IEEE/CVF international conference on computer vision},
  pages={16249--16258},
  year={2021}
}

@inproceedings{ryoo2023token,
  title={Token turing machines},
  author={Ryoo, Michael S and Gopalakrishnan, Keerthana and Kahatapitiya, Kumara and Xiao, Ted and Rao, Kanishka and Stone, Austin and Lu, Yao and Ibarz, Julian and Arnab, Anurag},
  booktitle={Proceedings of the IEEE/CVF conference on computer vision and pattern recognition},
  pages={19070--19081},
  year={2023}
}

@article{ryoo2024xgen,
  title={xgen-mm-vid (blip-3-video): You only need 32 tokens to represent a video even in vlms},
  author={Ryoo, Michael S and Zhou, Honglu and Kendre, Shrikant and Qin, Can and Xue, Le and Shu, Manli and Park, Jongwoo and Ranasinghe, Kanchana and Savarese, Silvio and Xu, Ran and others},
  journal={arXiv preprint arXiv:2410.16267},
  year={2024}
}

@inproceedings{xiong2024efficienttam,
  title={Efficient track anything},
  author={Xiong, Yunyang and Zhou, Chong and Xiang, Xiaoyu and Wu, Lemeng and Zhu, Chenchen and Liu, Zechun and Suri, Saksham and Varadarajan, Balakrishnan and Akula, Ramya and Iandola, Forrest and others},
  booktitle={Proceedings of the IEEE/CVF International Conference on Computer Vision},
  pages={11513--11524},
  year={2025}
}

@inproceedings{zhou2025edgetam,
  title={Edgetam: On-device track anything model},
  author={Zhou, Chong and Zhu, Chenchen and Xiong, Yunyang and Suri, Saksham and Xiao, Fanyi and Wu, Lemeng and Krishnamoorthi, Raghuraman and Dai, Bo and Loy, Chen Change and Chandra, Vikas and others},
  booktitle={Proceedings of the Computer Vision and Pattern Recognition Conference},
  pages={13832--13842},
  year={2025}
}
}

\end{document}